\documentclass[twocolumn,10pt]{asme2ej}

\usepackage{graphicx}
\usepackage{amsmath}
\usepackage{hyperref}   
\hypersetup{
	colorlinks=true,
	linkcolor=blue,
	citecolor=blue,
	urlcolor=blue,
}
\usepackage[square,numbers]{natbib}

\title{An In-Pipe Inspection Robot With Sensorless Underactuated Magnets and Omnidirectional Tracks: Design and Implementation}

\author{Gaurav Saha, K. M. Santosh, Anugu Reddy, Ravi Kant, Arjun Kumar}

\begin{document}

\maketitle    

This paper presents the plan of an in-pipe climbing robot that works utilizing an astute transmission part to investigate complex relationship of lines. Standard wheeled/proceeded in-pipe climbing robots are inclined to slip and take while investigating in pipe turns. The instrument helps in accomplishing the main inevitable result of getting out slip and drag in the robot tracks during advancement. The proposed transmission appreciates the practical furthest reaches of the standard two-yield transmission, which is developed the basic time for a transmission with three results. The instrument conclusively changes the track speeds of the robot considering the powers applied on each track inside the line relationship, by getting out the essential for any remarkable control. The amusement of the robot crossing in the line network in various orientation and in pipe-turns without slip shows the proposed game plan's adequacy.

\section{Introduction}

Pipe networks are unavoidable, in a general sense used to move fluids and gases in endeavors and metropolitan organizations. Most frequently, the lines are covered to agree with the security rules and to keep away from risks. This makes review and upkeep of lines truly testing. Covered lines are especially arranged to hindering, use, scale blueprint, and break commencement, accomplishing conveyances or harms that might affect egregious episodes. Different Inspection Robots \cite{han2016analysis} were proposed in the past to lead standard preventive appraisals to keep away from debacles. Moreover, bio-animated robots with crawler, inchworm, strolling parts \cite{choi2007pipe}, and screw-drive\cite{jalal2022pipe} frameworks were in like way demonstrated to be fitting for various necessities. In any case, the majority of them utilize dynamic controlling strategies to guide and move inside the line. Reliance upon the robot's course inside the line added to the difficulties, in like way leaving the robot frail against slip in the event that footing control techniques are not involved. The Theseus \cite{fujun2013modeling}, PipeTron \cite{debenest2014pipetron}, and PIRATE \cite{dertien2014design} robot series utilize separate portions for driving and driven modules that are interconnected by various linkage types. Each segments change or shift the bearing for getting sorted out turns. Also, great interesting controlling makes such robots dependent upon sensor information and significant assessment.

Pipe climbing robots with three changed modules are largely the more steady and give better portability. Prior proposed confined line climbers \cite{vadapalli2019modular,suryavanshi2020omnidirectional,vadapalli2021modular} have used three driving tracks coordinated equally to each other, as MRINSPECT series of robots. In such robots, to suitably control the three tracks, their rates were pre-depicted for the line turns. This addressed a basic for the robot to organize pipe-reshapes precisely at a specific bearing standing out from the pre-depicted rates \cite{vadapalli2019modular,suryavanshi2020omnidirectional}. In genuine applications, the robot's direction shift tolerating it encounters slip in the tracks during advancement. This restriction can be addressed by utilizing an inactively worked transmission part to control the robot. MRINSPECT-VI \cite{kim2016novel,kim2013pipe} utilizes a multi-fundamental transmission stuff part to control the speeds of the three modules. Regardless, for the division of the primary purpose and speed to the three modules, the arrangement of the transmission is utilized. This way of thinking made the central yield (Z) to turn speedier than the other two results (X and Y), making yield Z effectively impacted by slip \cite{kim2016novel}. This is caused considering the way that the outcomes of the transmission doesn't give equivalent energy to the information.

Other actually proposed manages genuine results with respect to transmissions \cite{kota1997systematic,ospina2020sensorless} also followed a similar plan. Transmission kills the alluded to constraint by perceiving vague result to enter dynamic relations \cite{vadapalli2021design}. In the envisioned plan every one of the three results are equivalently impacted by the data. This contributes for the robot to crash slip and drag toward any way of the robot during its turn of events. Also, the transmission part in the line climber upgrades the solace by reducing the reliance on the strong controls to go through the line affiliations. The framework precisely moves power and speed from a solitary obligation to the robot's three tracks through complex stuff trains, considering the stacks experienced by each track independently.
\section{Design}

A fanciful nonagon focal packaging of the robot houses three modules circumferentially $120^\circ$ segregated from one another. The transmission organized inside the focal occasion of the robot drives the three tracks by its driving sprockets through. slant outfits, that are connected with yields. The coordinated perspective on the instrument and is clear in the subsequent subsection. Every module houses a track and has openings for the four linkages to slide. The modules are pushed radially outwards with the assistance of direct springs mounted on the linkages (or shafts) i.e., the springs between the modules and the robot body. The fitting is a reference structure that restricts the advancement of the modules past OK endpoints. Right when the robot is sent inside the line, the spring-stacked tracks go through withdrew avoidance and presses against the line's inner dividers. It gives the principal footing to the robot to move. Every module in the robot can in like way pack unevenly, point by point in part. The tracks, obliged on the modules, are squeezed against the internal dividers of the line which give the huge equilibrium to the robot to move. There is a driving sprocket that changes over the rotational advancement from the eventual outcome of to translatory improvement for the robot. The divider squashing structure contains 4 direct linkages for the three modules in general. The modules have openings through which the straight linkages (or shafts) pass. They permit the advancement of the modules in the twisting headings from the turn of the robot.

\vspace{-0.15in}
\begin{figure}[ht!]
\centering
\includegraphics[width=2.5in]{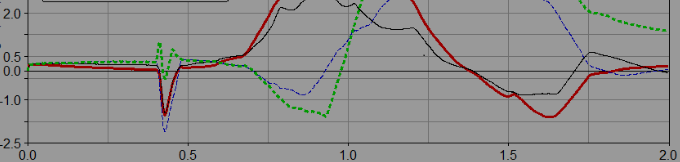}
\vspace{-0.15in}
\caption{\footnotesize Robot}
\label{1}
\vspace{-0.15in}
\end{figure}
\subsection{Transmission}

The transmission is the major constituent of the proposed robot. The instrument contains a solitary information, three transmissions, three two-input open transmissions and three results. The transmission's analysis is organized at the focal turn of the robot body. The three are composed consistently around the information, with a position of $120{^\circ}$ between any two. The are fitted radially in the middle. The single result of development the three results of the .Includes gear parts, for example, ring gears, slant cog wheels and side machine gear-piece wheels, while join ring gears, slant cog wheels and side pinion wheels. The side machine gear-piece wheels ofis agreed with their nearby side pinion wheels of, to move the power and speed from to it's bordering. The information of the worm gear give advancement to simuntaneously. Every two-yield transmission then, at that point, exchange got advancement to its lining two-input transmissions , reliant upon the heap conditions experienced by its different side pinion wheels. The improvement got by the side pinion wheels of additional makes an interpretation of them to the three results. The six transmissions and work together to unravel improvement from the obligation to the three results. Right when experience various loads, the side pinion wheels in exchanges various loads to the side pinion wheels of. Under this condition, makes a comprehension of transmission speed to its bordering. 

Precisely when experience equivalent weight or no stack, all of the side pinion wheels experience a tantamount weight making them turn at a relative speed and power. As fanned out, every one of the results share indistinguishable energy with the data. Furthermore, the results in like way share unclear energy with one another. This outcomes in the differentiation in loads in one of the results obliging an equivalent impact on the other two results that are undisturbed. The results work with indistinguishable paces when there is no heap or similar weight returning again to the results in general. The part works its results with transmission speed when the results are under changed loads. Precisely when one of the results is working at a substitute speed while the other two results are encountering tantamount weight, then, at that point, the two results with similar loads will work with indistinguishable paces. The part anticipated the robot, pushes its the three tracks with undefined velocities while moving inside a straight line. In any case, while moving inside the line twists, the transmission changes the track speed of the robot with a definitive objective that the track setting out to all parts of the more extended distance turns quicker than the track setting out to all parts of the more limited distance. Insinuate \cite{vadapalli2021modular,9635853} for extra figures, graphs and information.
\section{Mathematical Modeling Of The Robt}

\subsection{Design Optimisation}

The kinematic plot shows the openness of the affiliations and joints of the instrument. The transmission's kinematic and dynamic conditions are organized utilizing the bond diagram model. The information accurate speed from the engine is in like manner gave to the three ring gear-tooth wheels of the two-yield transmissions as. They turn at indistinguishable accurate speeds and with practically identical power for example times is the stuff degree of the obligation to the ring gears. Besides, a two-yield transmission organizes that the specific speed of its ring gear is the standard each period of the sassy speeds of its two side pinion wheels. These two side stuff tooth wheels can turn at various rates while remaining mindful of indistinguishable power \cite{deur2010modeling}.
\begin{equation}
f=12\mu kx,
\label{1}
\end{equation}
From the bond diagram model, we can construe that the three ring gears turns at indistinguishable accurate speeds and with practically identical power for example times the information running speed, occasions the information power, is the stuff degree of the obligation to the ring gears. Additionally, in a two-yield transmission, the specific speed of its ring gear is the conventional all of great importance of the sassy speeds of its two side pinion wheels. These two side machine gear-piece wheels can turn at various rates while remaining mindful of indistinguishable power \cite{vadapalli2021design}.The challenging rates are then inferred their side machine gear-piece wheels. Side machine gear-piece wheels are connected with a definitive objective that they don't have any broad improvement between the pinion wheels of a tantamount pair, i.e., precise speeds of side pinion wheels of an equivalent pair is generally indistinguishable. Subbing in (\ref{1}). Essentially, the result accurate speeds. The result to side stuff affiliation is diverged from accomplish the sprightly speed condition for the obligation to the results. The daring rates of the ring pinion wheels to the side machine gear-piece wheels, we accomplish an association among data and side gear-tooth wheels. Also, the outcome exact not entirely settled from the paces of the singular ring cog wheels to get an outcome to side stuff association. Comparing both the relations, we accomplish the exact speed condition for the commitment to the outcomes. where is the specific speed of the information is the stuff degree of the ring cog wheels to the result, while smart speeds of the result. The individual accurate speeds of the side stuff tooth wheels. Subsequently, the result accurate speeds are subject to the information running speed ($\omega_u$) and the side stuff yields. In the mean time, the powers of the ring gears is how much the powers of their seeing side pinion wheels. Like careful speed relationship for obligation to yields, by differentiating the ring pinion wheels with side pinion wheels relationship with the result to the ring gear affiliation, a relationship between yield powers is acquired.
\begin{equation}
TE=mg-12\mu (k.x)
\label{2}
\end{equation}
where is the specific speed of the information is the stuff degree of the ring cog wheels to the results, while and are running paces of the result. Yields are the individual accurate speeds of the side stuff tooth wheels. Thusly, the result accurate speeds are reliant upon the information buoyant speed and the side stuff yields. In the mean time, the powers of the ring gears is how much the powers of their seeing side pinion wheels. Like precise speed relationship for obligation to yields, by differentiating the ring pinion wheels with side pinion wheels relationship with the result to the ring gear affiliation, a relationship between yield powers to the information power is obtained.
\begin{equation}
N.C=0
\label{3}
\end{equation}
\begin{equation}
\tau = TE.r
\label{4}
\end{equation}
Conditions \eqref{3} and \eqref{4} presents the interest of the transmission to give equivalent advancement attributes in every one of the three results that encounters indistinguishable loads or when left unconstrained. Regardless, when the stuff parts experience a deterrent across an intersection point, the dapper rate and the power changes relying on the outside resistive power. The result speed of the transmission results are similar to the speed of the driving module sprockets.Therefore, the result speeds of the transmission are changed more than into track speeds. The information speed for the robot is, in this manner making a comprehension of to the results under practically identical stacking conditions. The sprocket broadness is steady for the three tracks in general. Ho Moon Kim et al. \cite{kim2013pipe}, in their paper proposed a procedure for working out the particular speeds of the three tracks inside pipe turns. Enduring that the robot enters the pipeline with the arrangement showed is the characteristic of the line turn, R is the expansiveness of musical development of the line twist and r is the extent of the line. Additionally, the speeds for their particular tracks B and C are gotten. The robot is embedded at various heading of the modules concerning OD. The changed paces for the not permanently set up for wind pipes in headings.
\begin{equation}
\tau = 0.38 N-m
\label{eqn01}
\end{equation}

\subsection{Constraints}

The straight springs in the module gives robot the adaptability to arrange turns with basically no issue. The most ideal strain in every module is $16 mm$. There are extra versatilities in the module openings, with the true that topsy turvy pressure is conceivable. This assists the robot with conquering obstacles and peculiarities in the line network that it might inspect veritable applications. The front consummation of the module is compacted totally anyway the back is in its most important extended state conceivable inside the line. The most silly unbalanced pressure permitted in a solitary module of the robot. In this way, $\phi$ is the best point the module can pack unevenly. Imply \cite{vadapalli2021modular,9635853} for extra figures, outlines and information.
\vspace{-0.1in}
\begin{figure}[ht!]
\vspace{-0.1in}
\centering
\includegraphics[width=2.75in]{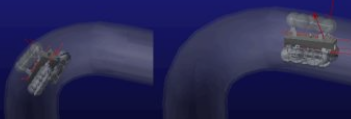}
\vspace{-0.12in}
\caption{\footnotesize Compression}
\label{asym(1)}
\vspace{-0.1in}
\end{figure}

\section{Simulations}

Redirections were coordinated to look at and support the improvement furthest reaches of the robot in different line affiliations. A tantamount will give us more experiences into the parts and direct of the made robot in genuine testing conditions. From this point forward, multi-body dynamic spreads was acted by changing over the course of action limits into an upgraded reenactment model. To diminish how much moving parts in the model and to diminish the computational weight, the tracks were altered into roller wheels. Every module houses three roller wheels in the improved model. Along these lines, the contact fix given by the tracks to the line dividers are decreased from 10 contact districts to 3 contact areas for every module. As far as possible prefer the track speeds and the module strain for each track A,B and C were dissected. Increases were performed by embedding the robot in three unquestionable headings of the module in both the straight lines and line turns. The robot is rehashed inside a line network organized by ASME B16.9 standard NPS 11 and plan 40. The redirections were composed for four assessment conditions in the line network containing Vertical locale, Elbow area, Horizontal piece and the U-part for various heading of the robot. The through and through distance of the line structure. The distance went by the robot in not permanently set up from point of combination of the robot body and the track's single distance still hanging out there from the middle roller wheel mounted in every module. Thusly, we get the out and out robot's direction, by eliminating the robot's length.The robot's way in vertical climbing and the last level locale is surveyed by deducting from their solitary piece length. The data of the is given a solid neat speed and improvement of the robot including the track speeds are considered in the reenactment.

In the upward piece and level area, the robot follows a straight way. Consequently, the tracks experience indistinguishable loads on every one of the three modules in both the examinations. Subsequently, the transmission gives indistinguishable speeds to the three tracks in general, like the robot's conventional speed. The saw track speeds in the redirection for the heading. Accordingly, every one of the qualities stand out from the hypothetical outcomes with a level out rate bungle (APE). This mistake surveys the genuine extent of deviation from the theoretical worth \cite{armstrong1992error}. To put together a key length of in vertical climbing, the robot anticipates that 0 should 9 seconds. the robot's capacity to climb the line facing gravity. Beginning the robot moves a distance of 350 mm in the boss even locale. In the more unassuming level piece of distance. In the elbow district, the robot moves at a steady distance to the place of assembly of bend of the line. The system changes the result paces of the tracks as shown by the parcel from the place of assembly of condition of the line. In all of the three headings of the robot, the external module tracks goes quicker to travel a more widened distance, while the interior module tracks turns even more postponed to set out to all parts of the most limited distance than the scope of twist of the line bend. In pipe reshapes, the diversion speeds of each track is displayed at the midpoint of seperately to brutal the saw track speeds without changes. The approximated speeds of each track is then stood apart from their particular theoretical rates with track down the all around rate bungle (APE).

For the course, the external modules (B and C) move at a normal speed of, while the interior module (A) move at a standard speed of 33.62mm/sec. These qualities interface with the hypothetical attributes. Also, the track speeds arranges with the ordinary worth of the expansion results. Moreover, the track speed an inspiration, diverge from the multiplication results with an APE of 2.5\%. Toward every way of both the straight and turn locales, the blunder respect is inconceivably irrelevant and they can be credited because of outside factors in clear conditions like crumbling. Accordingly, insignificant speed changes happens in the multiplication plot. From 9 to 24 seconds, the robot puts together the elbow area of distance, while it requires 59 seconds to go in the U-segment. The robot's capacity to investigate in pipe-turns. for shows that the external modules (B and C) move at the speed degree while the inside module (A) move in a speed degree. These attributes exist in the theoretical qualities. Also, the track speeds an inspiration, exist in the reenactment results. The speed range alluded to from the redirection is taken from the base to the most over the top summits of rates for each plots. Insinuate \cite{vadapalli2021modular,9635853} for extra figures, charts and information.

The expansion results for the track speeds in various headings, coordinates with the hypothetical outcomes got in locale. It is seen from the reenactment that in 60 seconds, the robot investigates all through the line network dependably at the embedded bearing. The outcome interface with the theoretical computation. This supports that the transmission disposes of slip and drag in the tracks of the line climber all over with no improvement misfortunes. In the proliferation, the robot is seen with no slip and drag all over, which further effects in lessened pressure impact on the robot and expanded advancement perfection. The track in the modules catch to the internal mass of the line to give balance during advancement. The springs are at first pre-stacked by a strain in all of the three modules similarly when embedded in the upward line area. In straight lines, the robot moves at the essential pre-stacked spring length. The distortion length increments by 1.5 mm for within and the external modules when the robot is moving close to elbow area and U-segment. This deformity clarifies the lengthy adaptability contemplated the modules to go through the creating cross-part of the line broadness in the twists during improvement.

\section{Conclusion}
The robot is given the brilliant transmission to control the robot unequivocally with no exceptional controls. The transmission has an undefined result to join energy, whose show is totally similar to the comfort of the standard two result transmission. The age results support strong crossing point of stunning line networks with twists of up to 180$^\circ$ in various headings without slip. Taking on the transmission part in the robot accomplishes the sharp postponed result of getting out the slip and drag all over of the robot during the turn of events. At the present, we are empowering a model to perform explores the proposed plan.
%
% ...

\bibliographystyle{asmems4}

\bibliography{asme2e}

\end{document}